\documentclass[10pt,twocolumn,letterpaper]{article}

\usepackage{cvpr}              

\usepackage{graphicx}
\usepackage{amsmath}
\usepackage{amssymb}
\usepackage{booktabs}
\usepackage{float}
\usepackage{dblfloatfix}
\usepackage{xcolor}

\usepackage[pagebackref,breaklinks,colorlinks]{hyperref}

\usepackage[capitalize]{cleveref}
\crefname{section}{Sec.}{Secs.}
\Crefname{section}{Section}{Sections}
\Crefname{table}{Table}{Tables}
\crefname{table}{Tab.}{Tabs.}

\newcommand{\grey}[1]{{\color{gray}{#1}}}
\newcommand{\smallgrey}[1]{{\footnotesize{\color{gray}{#1}}}}

\def\cvprPaperID{31} 
\def\confName{CVPR}
\def\confYear{2022}

\begin{document}

\title{Dynamic Query Selection for Fast Visual Perceiver}

\author{Corentin Dancette\\
Sorbonne Universit\'e\\
{\tt\small corentin.dancette@sorbonne-universite.fr}
\and
Matthieu Cord \\
Sorbonne Universit\'e, valeo.ai \\
{\tt\small matthieu.cord@sorbonne-universite.fr}
}
\maketitle

\begin{abstract}
Transformers have been matching deep convolutional networks for vision architectures in recent works.
Most work is focused on getting the best results on large-scale benchmarks, and scaling laws seem to be the most successful strategy: bigger models, more data, and longer training result in higher performance.
However, the reduction of network complexity and inference time remains under-explored.
The Perceiver model offers a solution to this problem: by first performing a Cross-attention with a fixed number $Q$ of latent query tokens, the complexity of the $L-$layers Transformer network that follows is bounded by $O(LQ^2)$.
In this work, we explore how to make Perceivers even more efficient, by reducing the number of queries Q during inference while limiting the accuracy drop.
\end{abstract}


\section{Introduction}
\label{sec:intro}

Convolutional neural networks have been the state-of-the-art in computer vision for several years~\cite{Krizhevsky2012alexnet, resnet2016}. However, Transformer architectures, originally designed for natural language processing~\cite{vaswani2017attention} have been matching deep convolutional networks for vision tasks in recent works~\cite{dosovitskiy2021vit, touvron2021training, touvron2021cait, liu2021swin, d2021convit}.
Scaling is one of the most successful strategies to get better results on large-scale benchmarks: bigger models, more data, and longer training result in higher performance \cite{dosovitskiy2021vit}. 
However, the main drawback of the transformer architecture is its quadratic complexity in the number of tokens.
For large inputs, such as very big images or texts, using a regular transformer is not possible. Thus, making transformers more efficient and breaking their quadratic complexity bottleneck has important practical uses, as it would allow using them on larger inputs and running those architectures on more devices. Many works propose approximations of the attention mechanism to reduce its complexity~\cite{wang2020linformer, kitaev2020reformer, xiong2021nystromformer, beltagy2020longformer,  beltagy2020longformer}.
The Perceiver~\cite{jaegle2021perceiver, jaegle2021perceiverio} model introduces a different solution to this problem: it reduces the number of tokens by first performing Cross-Attention with a fixed number of latent queries Q, which results in a fixed number of Q tokens. The complexity of each transformer Self-Attention layer that follows is bounded by $O(Q^2)$.

We propose to go further to speed up the Perceiver by exploring how to consider for each image only a subset of $K$ queries among the $Q$, with a performance drop as small as possible. Overall, this can reduce the number of queries at inference, allowing faster predictions. This may be very useful if a ``quick-and-dirty'' answer is required at times, and at other times, more accurate answers are required.
Naive query subsampling strategies in a regular Perceiver do not work: reducing the number of queries randomly decreases significantly the accuracy. Instead, we propose (a) a learning strategy to minimize the accuracy drop, and evaluate its impact experimentally, and (b) a method to dynamically select queries to be removed to minimize the performance drop.


\section{Perceiver Speed-up Method}
\paragraph{Visual Perceiver architecture}

\begin{figure*}[h]
    \centering
    \includegraphics[width=1.0\textwidth]{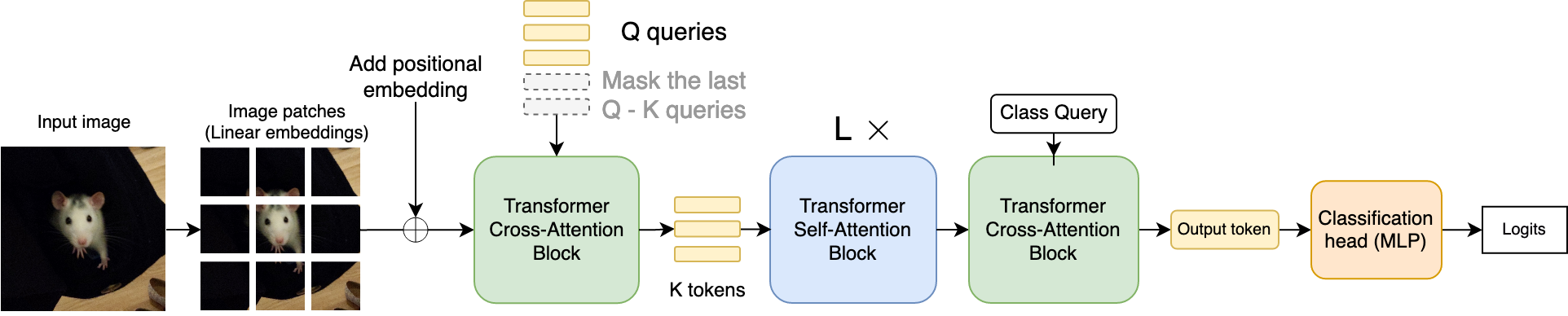}
    \caption{Our VisualPerceiver architecture. The input image is divided into patches and fed into a linear layer. A Cross-Attention block is applied using Q latent queries. Finally, a regular transformer architecture is applied, with a final decoder and a classification head. }
    \label{fig:model}
\end{figure*}

We start from the original Perceiver-IO\cite{jaegle2021perceiverio} architecture with some modifications designed for the image classification task. The Perceiver model is based on the Transformer\cite{vaswani2017attention} architecture. In Perceiver-IO, the input is first split into $N$ tokens. For images more specifically, each pixel is used as a single token. This results in a large number of tokens (e.g. $50$K for $224$x$224$ images), which cannot all be processed by a Transformer, due to its quadratic complexity in the number of tokens. The Perceiver reduces this number of tokens to a fixed number $Q$ by using a Cross-Attention layer.
The input tokens are fed to the cross-attention block as keys and values, and there are $Q$ query tokens initialized randomly and learned during training. Each query token will attend to various input tokens and aggregates the result into an output token.
Then, a regular Transformer encoder is applied to those $Q$ tokens (a sequence of multiple Self-Attention blocks), and at the end, a Transformer-Decoder layer is used to return with a single token. A classification head then takes this token and returns output logits.

We make the following changes to the original Perceiver-IO architecture:
\textit{First}, as shown in Figure~\ref{fig:model}, we use an image processing similar to ViT\cite{dosovitskiy2021vit}: instead of using one token per pixel as in the original Perceiver-IO, we use image patches. For CIFAR-10, we use 4x4 patches. This reduces the number of tokens by a factor of 16 and gives us a total of 64 tokens. This adds a useful prior and reduces the computation in the network.

\textit{Then}, we use learned positional embeddings instead of fixed Fourier embeddings. As explained in \cite{jaegle2021perceiver, jaegle2021perceiverio}, learned positional embeddings did not work very well for images in the original Perceiver, probably due to the important number of tokens. In our case, with patch tokens, the model is able to learn useful positional embeddings.

\vspace{-2mm}
\paragraph{Query Masking}

Our objective is a model where the user can choose the complexity of the model by selecting an arbitrary number of queries. If we remove random queries from a Visual Perceiver trained with a fixed number of queries, the accuracy drops very quickly, as we show in Section~\ref{sec:main-experiments}. We could also train one model for each number of queries, but this is extremely costly in training time and storage.
To solve this issue, we propose the \textbf{Query Masking} training strategy. We train a single Visual Perceiver that is initialized with a fixed number of queries Q. During training, for each batch, we first select a random number of queries $K \in [1, Q]$. For the whole batch, we use only the K first queries in the Cross-Attention layer, as shown in Figure~\ref{fig:model}. Note that the queries are ordered: for a given K, the same K queries will always be selected. This means that the first query is used in all batches, while the last query is only used in approximately every 1/Q batch.
For evaluation, we use the same number of queries K for the whole dataset. This gives us an accuracy score for each K.

Additionally, in Section~\ref{sec:dqs}, we propose a Dynamic Query Selection method, to adaptatively select this number K for each example to make the inference even faster without accuracy loss

\section{Experiments}
\label{sec:experiments}
We use the CIFAR-10 dataset for our experiments. We divide the image into 64 patches of size 4x4 pixels. Each patch is projected in a 192-dimension embedding. Our transformer architecture is based on the DeIT-Tiny model: it has 12 Self-Attention layers with 3 heads. The dimension of all tokens is 192 throughout the network. The model has 6.18 M parameters.
We train all models with the same training procedure as DeIT~\cite{touvron2021training} for 350,000 steps, with a learning rate of 5.e-4 and a batch size of 512.

\subsection{Main experiment}
\label{sec:main-experiments}
\begin{figure}[h]
    \centering
    \includegraphics[width=\linewidth]{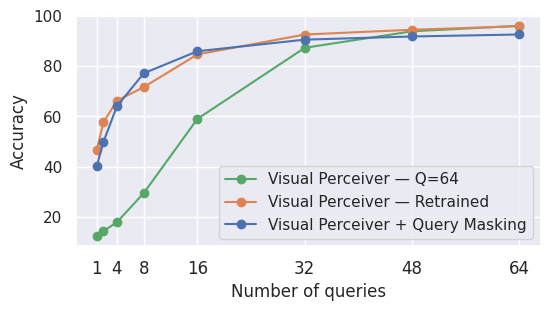}
    \caption{Accuracies for various tokens during inference, for three types of models. In blue, our Visual Perceiver, which was trained with a various number of tokens. In orange, is the Perceiver, with a specific model trained for each different number of tokens. In green, is the original Perceiver, which was trained only once with the maximum number of tokens.}
    \label{fig:chart1}
\end{figure}

We compare our strategy with a baseline and a topline on CIFAR-10.
As \textbf{baseline}, we train a model with a fixed number of 64 tokens, the same as the number of input patches. We note it \textit{Visual Perceiver - Q=64}.
We then evaluate this model with various numbers of tokens. For each number K, we run the evaluation 5 times, with each time K random tokens sampled from the 64 tokens.
The results are reported in Figure~\ref{fig:chart1} in green.
We note that the accuracy decreases rapidly when we move away from the 64 tokens used in training.

For the \textbf{topline}, we train specialized Visual Perceiver models for each number of queries
As we see in Figure~\ref{fig:chart1} in orange, the models perform much better than the baseline Perceiver trained only with 64 tokens with naive query subsampling.
Our goal is to stay as close as possible to the performances of this model while training a single model.

The results for our \textbf{Visual Perceiver with Query Masking} are reported in Figure~\ref{fig:chart1} in blue: we see that the accuracy is very close compared to the retrained specialized models while using a single model for all evaluations.

\subsection{Efficiency}

We report in Table~\ref{tab:efficiency} the number of Floating-Point operations and the inference time for our Visual Perceiver model. The inference time is computed for a batch of 512 images.  The model's inference time is reduced by more than half by using 32 queries instead of 64, with a small accuracy loss. We display the accuracies of those three models with respect to the FLOP and inference time in Figure~\ref{fig:chart-flops} of the Appendix.

\begin{table}[h]
    \centering
    \begin{tabular}{rccc}
    \toprule
    \# Q & FLOP (M) & Inference time (ms) & Acc. \\
    \midrule
    1 & 11 & \phantom{0}38.34\phantom{--} \smallgrey{(0.05)} & 40.29\\
    2 & 17 & \phantom{0}47.81\phantom{--} \smallgrey{(0.06)} & 49.89 \\
    4 & 29 & \phantom{0}57.68\phantom{--} \smallgrey{(0.08)} & 64.00 \\
    8 & 52 & 91.45\phantom{--} \smallgrey{(0.12)} & 77.13\\
    16 & 99 & 169.77\phantom{--} \smallgrey{(0.23)} & 85.91 \\
    32 & 195 & 330.74\phantom{--} \smallgrey{(0.46)} & 90.52 \\
    48 & 293 & 512.62\phantom{--} \smallgrey{(0.72)} & 91.78 \\
    64 & 394 & 712.08\phantom{--} \smallgrey{(\textbf{1.00})} & 92.58 \\
    \bottomrule
    \end{tabular}
    \caption{Floating-Point operations and inference time for a batch of 512 images for various numbers of queries for the Visual Perceiver model. The numbers in \grey{brackets} are relative inference times compared to the full model with 64 queries. \vspace{-3mm}}
    \label{tab:efficiency}
\end{table}

\subsection{Analysis of the Perceiver Cross-attention}

\begin{figure}[h]
    \centering
    \includegraphics[width=0.9\linewidth]{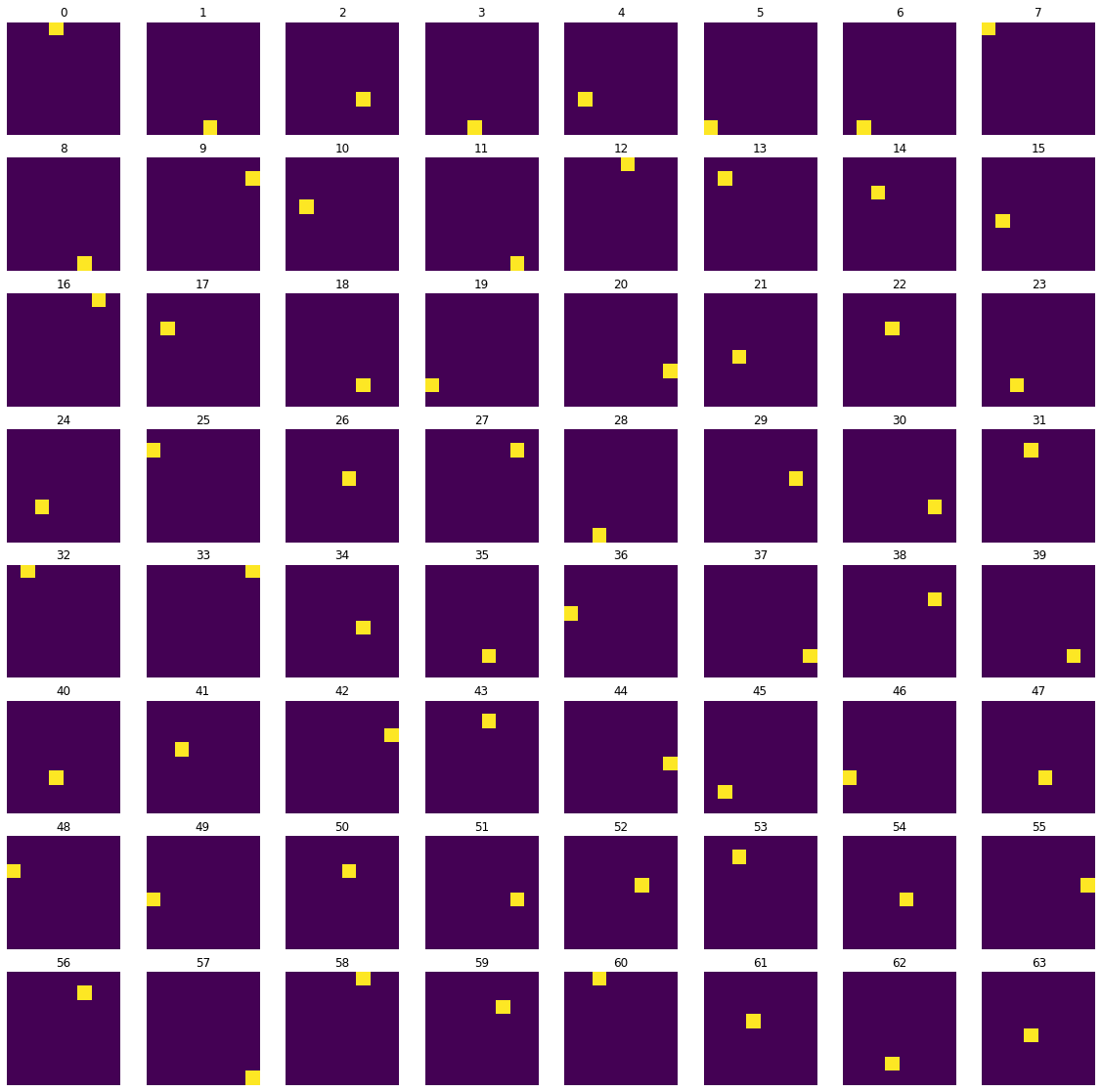}
    \caption{Average attention for the 64 queries in the first Cross-Attention for the Visual Perceiver model trained with a fixed Q=64. All queries are perfectly localized in the image: they select a single input patch token.} \label{fig:regular-perceiver-64}
\end{figure}

\begin{figure}[h]
    \centering
    \includegraphics[width=0.7\linewidth]{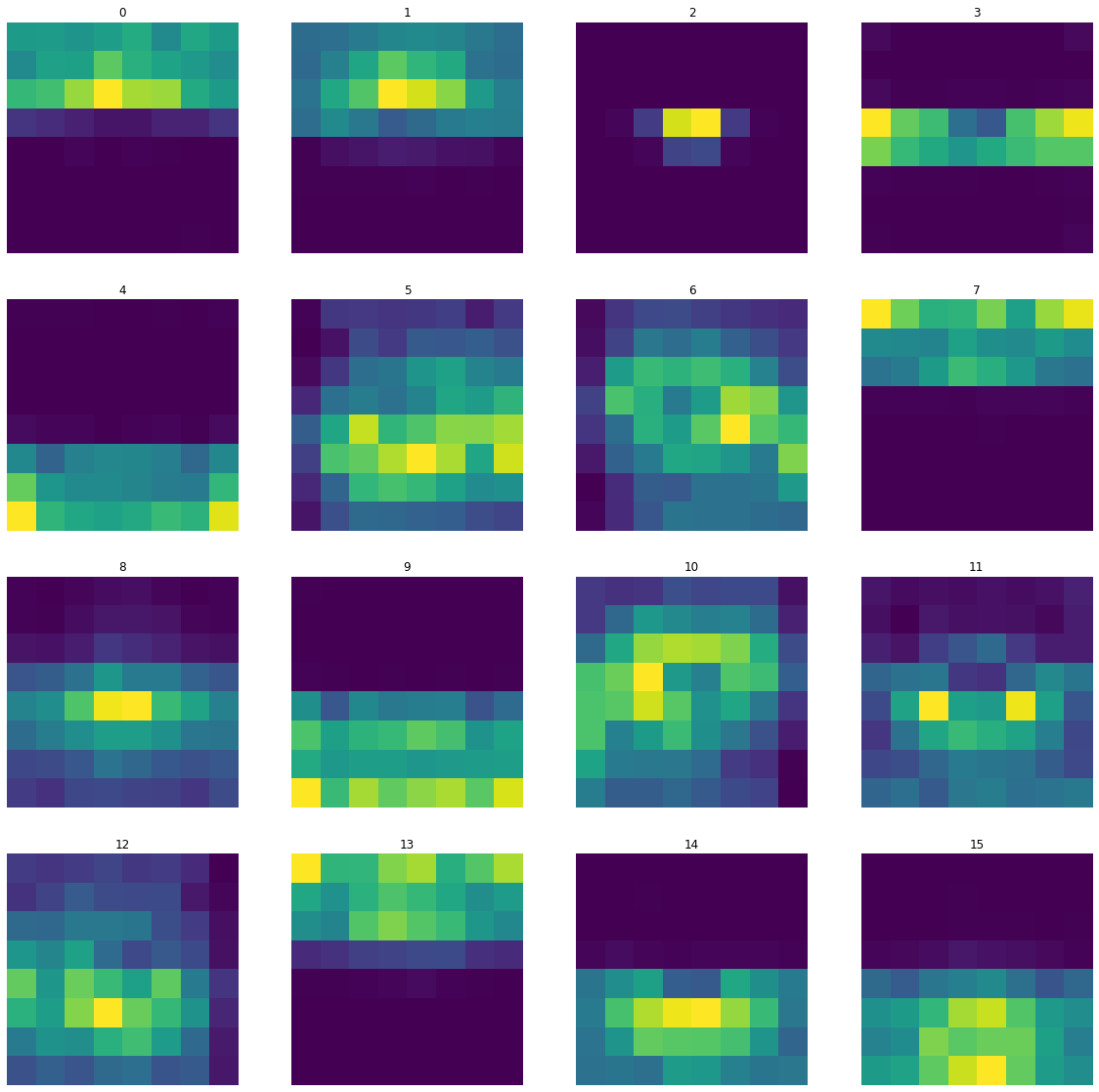}
    \caption{Average attention maps for the 16 queries for the first Cross-Attention in the Visual Perceiver trained with a fixed Q=16. We see that each query focuses on regions instead of single pixels, as the model is limited to a small number of queries.}
    \label{fig:regular-perceiver-16}
\end{figure}

We report here the average attention maps of the Cross-Attention layer for various models. All attention maps are averaged across the CIFAR-10 validation set. We display a single attention map for every query, as the Cross-Attention layer has a single head.
First, we display in Figure~\ref{fig:regular-perceiver-64} the attention maps for a Visual Perceiver trained with the full 64 queries during the whole training (without any query masking). Each 8x8 square corresponds to a single query, and the colors correspond to the attention score between this query and a specific image patch, averaged across the dataset.
We see that each query is perfectly aligned with a single image patch. This means that each Visual Perceiver learned to align the queries with the learned positional encodings, and the resulting model is very similar to a regular Vision Transformer.
In Figure~\ref{fig:regular-perceiver-16}, we display a similar series of attention maps, for a Visual Perceiver trained with fixed Q=16 queries. We see that each query attends to large regions of the input image. We display additional attention maps in the Appendix.

\begin{figure}[h]
    \centering
    \includegraphics[width=0.9\linewidth]{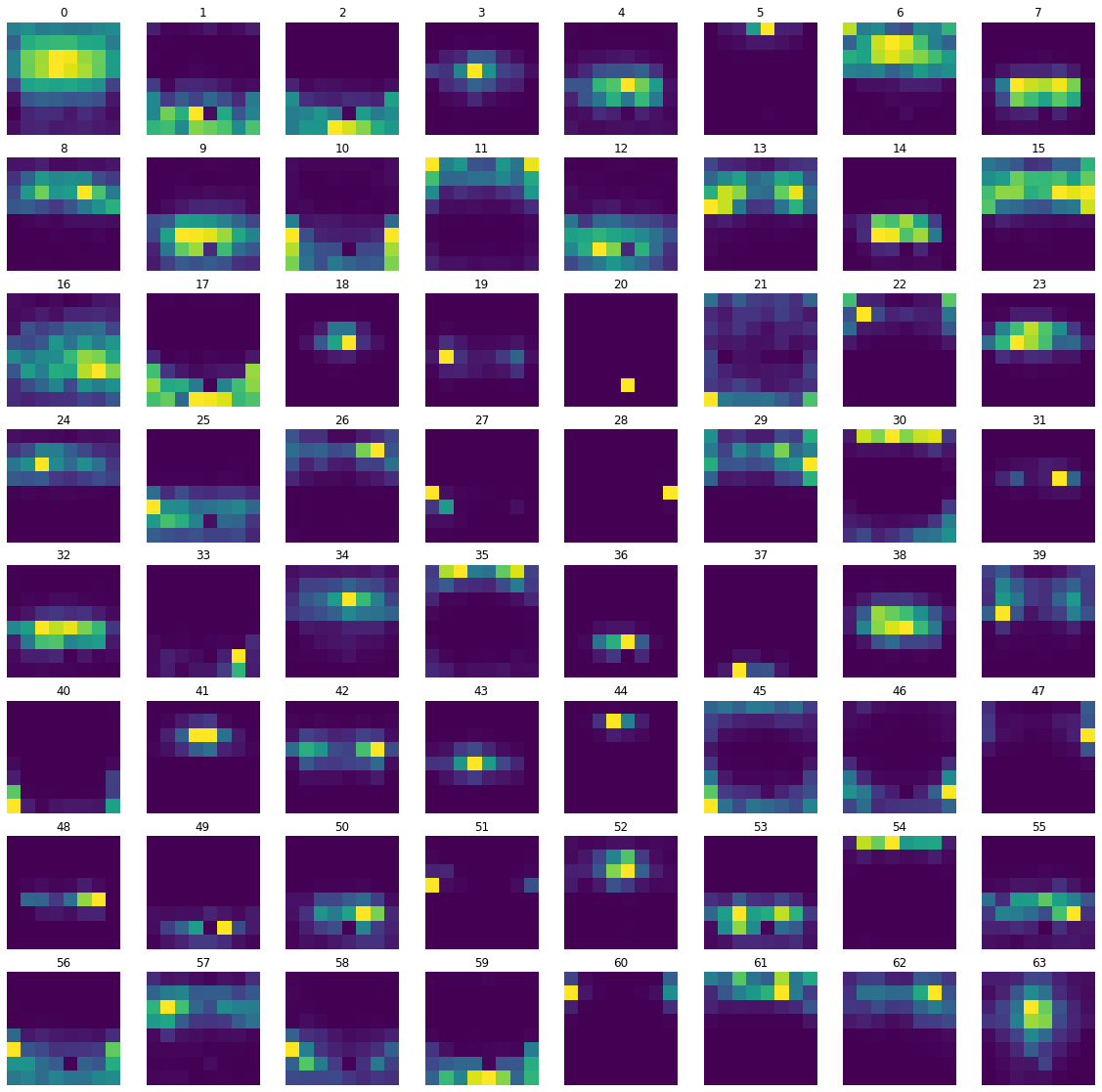}
    \caption{Average attention maps of the 64 queries in the first Cross-Attention for our Visual Perceiver + Query Masking.
    The first query attends to a large part of the image, while the following queries are localized in smaller regions.\vspace{-2mm}
}
    \label{fig:perceiver-masking}
\end{figure}

Then, Figure~\ref{fig:perceiver-masking} shows the attention map for our Visual Perceiver trained with Query Masking. During training, we sample the first K queries, with a K sampled uniformly between 1 and 64. The queries are ordered from left to right, then up to bottom. We see that the attention maps are extremely different from the regular Visual Perceiver maps in Figure~\ref{fig:regular-perceiver-64}. The first query has learned to attend to a large portion of the image, with a slightly larger weight in the middle, where the objects are usually located. This is unsurprising, as the model is required to predict the answer using only the first query, so it needs to aggregate as much information as possible. Subsequent queries specialize in smaller regions.

\subsection{Dynamic Query Selection}
\label{sec:dqs}

We showed that we can reduce the number of queries significantly with a small performance drop. Here, we investigate if we can find a better balance between compute time and accuracy. As more complex images might need more tokens than simple images, we try to mitigate the performance drop by selecting dynamically for each example, in inference only, a specific number of queries. We call this strategy Dynamic Query Selection (DQS).
As a reminder, in the Cross-Attention block: The $n_x$ input tokens $X$ are projected to key tokens $K$ and value tokens $V$. We also have the $n_q$ query tokens $Q$, which are parameters of the network in our case. The $n_q$ output tokens are $Y = \mathrm{softmax}\left(\frac{Q K^{T}}{\sqrt{d_{k}}}\right)V$ where $d_k$ is a scaling factor. Each query $q_i$ gives an output token $y_i$, which is a weighted average of all value tokens $v_j \in V$. 

We use the following procedure:
We compute pairwise cosine similarities between the output tokens: $s_{i,j} = \cos(y_i, y_j)$ where $(i, j) \in [1, n_q]^2$.
Then, we remove queries that have a cosine similarity with one of the previous tokens higher than a threshold $t$.
If two queries aggregate very similar image regions, their representation will be very close, and one of the two tokens will be discarded.
We display an illustration of this strategy in Figure~\ref{fig:dynamic-query-selection} of the Appendix.

\begin{figure}[h]
    \centering
    \includegraphics[width=\linewidth]{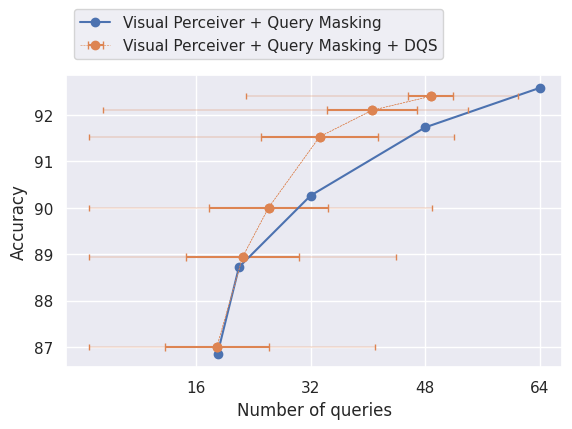}
    \caption{Our dynamic query selection. 
    For the model with DQS, the number of queries displayed on the horizontal axis is the average over the whole dataset. The error bars in bold show the standard deviation of the number of queries, while the thin error bars show the min and max number of queries.\vspace{-2mm}}
    \label{fig:res-dynamic-selection}
\end{figure}

\paragraph{Results}
The results are displayed in Figure~\ref{fig:res-dynamic-selection}. We show our strategy in orange, with various thresholds $t$.
We show the accuracy with respect to the average number of tokens used in the validation set. This makes it possible to compare fairly our Visual Perceivers with regular Query Masking, and our model with DQS.
For the same number of queries, i.e. the same average computation budget, we can see that the model with DQS performs significantly better than the fixed query selection. For example, for a desired accuracy of  $\sim$91.5, the regular Visual Perceiver with Query Masking requires about 48 queries, while our model has an average number of queries of 32, reducing significantly the average inference time. And if we use an average of 48 queries, our model's accuracy with DQS goes up to 92.4.
This confirms the effectiveness of our method to reduce inference time even further without sacrificing accuracy.



\vspace{-3mm}
\paragraph{Conclusion}
We proposed a variation of the Perceiver architecture designed for Image Classification, and a Query-Masking learning strategy to enable the model to work with a various number of tokens. This makes it possible to dynamically choose a trade-off between compute time and accuracy, which is useful in situations where a fast answer is required.
This opens the perspective for future work: First, this method could be tested on bigger images and datasets, for example on the Imagenet dataset~\cite{imagenet}. This approach could bring bigger improvements in inference time for larger images.
Then, there is a need for better methods to select the number of queries dynamically, based on the complexity of the image. This would allow a model to further reduce its inference time without any intervention from the user, and without sacrificing the accuracy.




\paragraph{Acknowledgments}
This work was granted access to the HPC resources of IDRIS under the allocation 20XX-AD011011588R2 made by GENCI.

{
\small
\bibliographystyle{ieee_fullname}
\bibliography{egbib}
}

\clearpage
\section{Appendix}

\subsection{Additional results}

\begin{figure}[h]
    \centering
    \includegraphics[width=\linewidth]{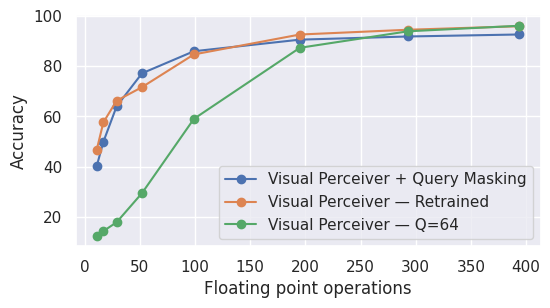}
    \caption{Accuracies for our three models with respect to their FLOP count.}
    \label{fig:chart-flops}
\end{figure}

\begin{figure}[h]
    \centering
    \includegraphics[width=\linewidth]{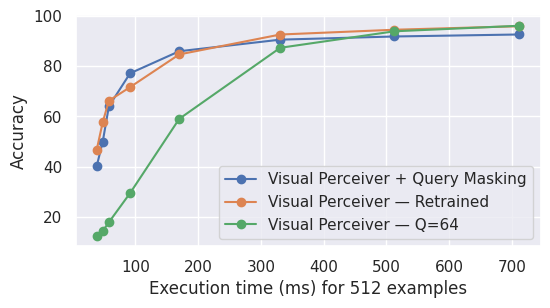}
    \caption{Accuracies for our three models with respect to their inference time.}
    \label{fig:chart-inferencetime}
\end{figure}

We display in Figure~\ref{fig:chart-flops} and Figure~\ref{fig:chart-inferencetime} the same results as in Figure~\ref{fig:chart1} from the main paper, with respectively the floating-point operations and the inference time on the horizontal axis. We see that the reduction of the number of queries results directly in a reduction in the number of floating-point operations and a faster inference.

\paragraph{Table of results} Additionally, we display in Table~\ref{tab:results-main-fig} and Table~\ref{tab:results-fig-dqs} the numerical results that were used to plot Figure~\ref{fig:chart1} and ~\ref{fig:res-dynamic-selection} respectively.

\begin{table}[h]
    \centering
    \begin{tabular}{ccccc}
    \toprule
            & \multicolumn{3}{c}{Accuracy of Visual Perceiver} \\
    \cmidrule{2-4} 
     \# Q & Q. Masking & Retrained & Q=64 \\
     \midrule
       1  & 40.29 & 46.78 & 12.51 \\
       2  & 49.890 & 57.63 & 14.36 \\
       4  & 64.00 & 66.23 & 17.87 \\
       8  & 77.13 & 71.63 & 29.47 \\
       16 & 85.91 & 84.64 & 59.01 \\
       32 & 90.52 & 92.56 & 87.28 \\
       48 & 91.78 & 94.46 & 93.84 \\
       64 & 92.58 & 95.87 & 96.05 \\
     \bottomrule
    \end{tabular}
    \caption{Numerical results for Figure \ref{fig:chart1}.  ``Q. Masking'' stands for Query Masking.}
    \label{tab:results-main-fig}
\end{table}

\begin{table}[h]
    \centering
    \begin{tabular}{ccc}
    \toprule
    Average Q & Threshold $t$ & Accuracy \\
     \midrule
    18.92 & 0.6\phantom{0} &  87.01 \\
    22.51 & 0.65 & 88.94 \\
    26.13 & 0.7\phantom{0} &  90.00 \\
    33.27 & 0.8\phantom{0} & 91.53 \\
    40.62 & 0.9\phantom{0} & 92.10 \\
    48.77 & 0.99 & 92.40 \\
     \bottomrule
    \end{tabular}
    \caption{Numerical results for the Visual Perceiver + Query Masking + Dynamic Query Selection from Figure~\ref{fig:res-dynamic-selection}}
    \label{tab:results-fig-dqs}
\end{table}

\subsection{Additional attention maps}

\begin{figure}[h]
    \centering
    \includegraphics[width=0.3\linewidth]{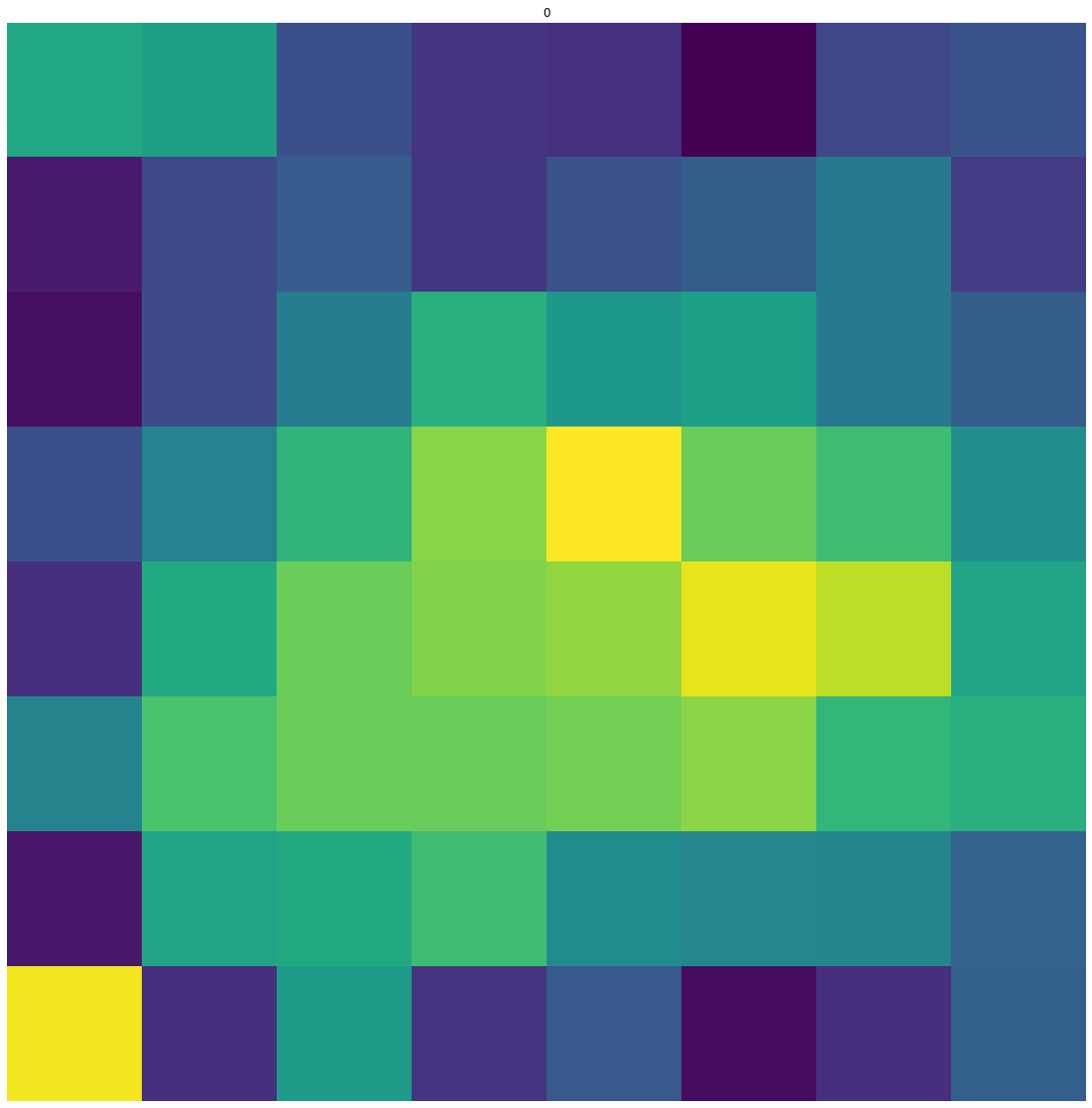}
    \caption{Attention maps for the query for the first Cross-Attention layer for the regular VisualPerceiver model trained with 1 query. We note that this query does not attend to a specific part of the image, but averages a large part of the image. This is similar to the first token in our Visual Perceiver model.}    \label{fig:regular-perceiver-1}
\end{figure}

\paragraph{Visual Perceiver - 1 Fixed Query}
In Figure~\ref{fig:regular-perceiver-1}, we display the attention for a Visual Perceiver trained with a single query. We see that the attention pattern is very similar to the first query from our Visual Perceiver trained with a Query-Masking strategy: it attends to a large region of the image, centered in the middle, where the objects are usually located. This is expected, as the model needs to aggregate as much information as possible in a single token.

\subsection{Visualization of the Query Selection Strategy}

\begin{figure}[h]
    \centering
    \includegraphics[width=0.8\linewidth]{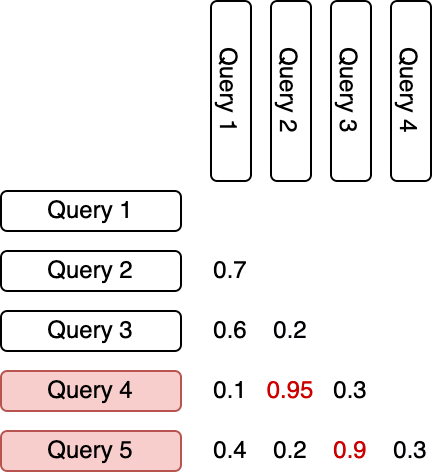}
    \caption{Illustration of the dynamic query selection. Each score is the cosine similarity between two tokens. With a threshold of $t=0.9$, the last two queries in red will be removed, as their similarity with one of the previous queries is higher than 0.9.}
    \label{fig:dynamic-query-selection}
\end{figure}

In Figure~\ref{fig:dynamic-query-selection}, we display an illustration of the dynamic query selection strategy. This allows us to remove dynamically useless queries for each example, without having to define a fixed number of queries K common for all examples.

\end{document}